\pdfoutput=1

\documentclass[11pt]{article}

\usepackage[final]{acl}

\usepackage{times}
\usepackage{latexsym}

\usepackage[T1]{fontenc}

\usepackage[utf8]{inputenc}

\usepackage{microtype}

\usepackage{inconsolata}

\usepackage{graphicx}
\usepackage{listings}
\usepackage{xcolor}
\usepackage{comment}
\usepackage{fancyvrb}
\usepackage{adjustbox}
\usepackage{wrapfig}
\usepackage{url}
\usepackage{booktabs}
\usepackage{caption}
\usepackage{hyperref}

\usepackage[utf8]{inputenc}
\usepackage[T5,T1]{fontenc}
\usepackage[english]{babel}

\title{Evaluating Multilingual Long-Context Models for Retrieval and Reasoning}

\author{Ameeta Agrawal, Andy Dang, Sina Bagheri Nezhad, Rhitabrat Pokharel, \\
        \textbf{Russell Scheinberg}\\
        Department of Computer Science\\
        Portland State University, USA\\
        \texttt{\{ameeta, andang, sina5, pokharel, rschein2\}@pdx.edu}
        }

\begin{document}
\maketitle
\begin{abstract}
Recent large language models (LLMs) demonstrate impressive capabilities in handling long contexts, some exhibiting near-perfect recall on synthetic retrieval tasks. However, these evaluations have mainly focused on English text and involved a single target sentence within lengthy contexts. Our work investigates how LLM performance generalizes to multilingual settings with multiple hidden target sentences. We create a new dataset -- \texttt{mLongRR} -- to comprehensively evaluate several multilingual long-context LLMs on retrieval and reasoning tasks across five languages: English, Vietnamese, Indonesian, Swahili, and Somali. These languages share the Latin script but belong to distinct language families and resource levels. Our analysis reveals a significant performance gap between languages. The best-performing models such as Gemini-1.5 and GPT-4o, achieve around 96\% accuracy in English to around 36\% in Somali with a single target sentence. However, this accuracy drops to 40\% in English and 0\% in Somali when dealing with three target sentences. Our findings highlight the challenges long-context LLMs face when processing longer contexts, an increase in the number of target sentences, or languages of lower resource levels. 
\end{abstract}

\section{Introduction}


The ability to model long context sequences spanning tens of thousands of tokens is crucial for tasks such as summarization and question answering based on long documents such as books and reports, and code generation at the repository level. Recent advancements in large language models (LLMs) have focused on improving their capabilities in processing long context information \citep{dai2019transformer, chen2023extending,ding2024longrope}.


Long-context language models, particularly multilingual ones, have the potential to enable remarkable progress in various applications by  understanding lengthy textual data across different languages. An example of this potential was recently demonstrated by the newly introduced Gemini-1.5 Pro model \citep{reid2024gemini} which leveraged its long-context window for in-context learning. By including a grammar manual in its context window, the model was able to learn to translate from English to Kalamang, an extremely low-resource language with fewer than 200 speakers \citep{visser2020kalamang}. Such examples highlight the potential of long-context models in tackling challenging tasks in low-resource languages, where data scarcity has traditionally been a barrier.

Current methods for evaluating long-context LLMs primarily focus on English text. This has led to a severe lack of insights into their performance across diverse languages. Evaluating multilingual performance is crucial, not only for informing the development of effective models that serve diverse communities \cite{lai-etal-2023-chatgpt, ahuja-etal-2023-mega}, but also for developing safer models as research suggests that LLMs tend to generate more unsafe and irrelevant responses to malicious prompts in lower-resource languages \citep{shen2024language}. However, there is a notable lack of multilingual benchmarks hindering our understanding of how long-context LLMs perform across different linguistic contexts.

\begin{table*}[t]
    \centering
    \begin{tabular}{l|c|c|l|c} 
    \toprule
    \textbf{Language} & \textbf{ISO 639-3 Code} & \textbf{Resource Level} & \textbf{Language Family}    & \textbf{Script}  \\ 
    \midrule
    English          & \texttt{eng}           & Level 5                & Indo-European   & Latin      \\ 
    Vietnamese       & \texttt{vie}           & Level 4                & Austro-Asiatic    & Latin    \\ 
    Indonesian       & \texttt{ind}           & Level 3                & Austronesian  & Latin       \\ 
    Swahili          & \texttt{swa}           & Level 2                & Niger-Congo      & Latin          \\ 
    Somali           & \texttt{som}           & Level 1                & Afro-Asiatic    & Latin      \\ 
\bottomrule
    \end{tabular}
    \caption{Languages studied and their details.}
    \label{tab:language_resources}
    
\end{table*}

To address this gap, we present the first comprehensive study of long-context LLMs in multilingual settings leveraging evaluation frameworks relying on synthetic tasks  \citep{mohtashami2023landmark,chen2023extending,liu2024lost,gkamradt,reid2024gemini,claude3}. Although the task is partially synthetic, we create a new dataset -- \texttt{mLongRR}\footnote{The code is available at \url{https://github.com/PortNLP/mLongRR}.} -- consisting of naturally occurring text and human translated data, making the setup as close to real-world setting while creating a controlled environment for comparing model performance across languages. In addition to retrieval tasks, we introduce a new reasoning task where the models not only need to retrieve relevant items but also compare them with each other. For this, the models must keep track of these items in a long-context scenario, allowing us to analyze the models' reasoning capabilities.

We conduct a systematic evaluation of six different LLMs across five languages of varying resource levels. Our research aims to answer the following two questions: 

\noindent (1) How do the long context capabilities of LLMs compare in retrieval and reasoning tasks in multilingual contexts? 

\noindent (2) Are there significant performance differences between LLMs in multilingual contexts?

Some of our key findings are summarized as follows: 
\begin{itemize}
\item The performance rapidly declines as we increase the context lengths for all languages.
\item The performance also rapidly decreases as we move from higher-resource to lower-resource languages.
\item Reasoning tasks are more challenging than retrieval tasks for all languages.
\item There is a significant gap between the performance of different LLMs.
\item Even seemingly simple ``needle in the haystack'' evaluation is able to  expose limitations in current models when dealing with multilingual contexts.
\end{itemize}

We hope that the findings of our study will contribute to a deeper understanding of current long-context evaluation in multilingual contexts and encourage the development of more effective long-context multilingual models.

\section{Related Work}

Recent advancements in language models have focused on improving their ability to recall and reason over fine-grained information from tens of thousands of tokens of context \citep{achiam2023gpt, jin2024llm}. Due to shortage of really long-context benchmarks, evaluation is typically focused on synthetic tasks such as passkey retrieval \citep{mohtashami2023landmark,chen2023extending,liu2024lost,ding2024longrope,jin2024llm} or needle in a haystack \citep{gkamradt,reid2024gemini,claude3} which measure a model's ability to accurately recall information from a vast corpus of data.

Recently, Gemini 1.5 \citep{reid2024gemini} and Claude-3 \citep{claude3} models reported near-perfect recall on the needle in a haystack task. Prior work has also studied perplexity but a low perplexity score has shown to not necessarily indicate proficiency in handling long contexts or reflect the model’s performance on sequence-level tasks in real applications \citep{sun2021long,pal2023giraffe,jin2024llm}. Furthermore, most of these studies have been limited to English only texts.

Although some long-context real-world benchmarks have been recently introduced, they are also limited to English \citep{an2023eval}, and while some bilingual English/Chinese \citep{bai2023longbench,qiu2024clongeval,yuan2024lveval} datasets offer a slight improvement, due to the effort-intensive nature of dataset creation, they are limited to a very small number of languages.





\section{Multilingual Needles in a Haystack for Retrieval and Reasoning Evaluation}

\subsection{Languages and Models}

\noindent \textbf{Languages} \quad We selected five languages to study: English, Vietnamese, Indonesian, Swahili, and Somali. These languages span  different resource levels from high to extremely low\footnote{The linguistic diversity taxonomy \citep{joshi-etal-2020-state} is used to identify the resource levels.} allowing us to gain insight into how language resource levels affect models' ability to work over long context windows.

Furthermore, we deliberately control for script-related variables as they have been shown to have a considerable impact on the performance of a model \cite{bagheri-nezhad-agrawal-2024-drives}. We study languages that use the Latin script for three reasons: models perform significantly better on Latin-script languages than non-Latin languages  \citep{chau2021specializing,bang2023multitask}, the fragmentation rate of Latin script is lower than other scripts allowing Latin-script languages to be represented with substantially fewer tokens as compared to languages in other scripts \citep{acs,ahiaetal2023languages} -- a disparity that becomes even more pronounced over long contexts, and, lastly, the fragmentation rate of Latin-script languages remains comparable which is helpful when considering considerably long input texts. Our selection of languages, shown in Table \ref{tab:language_resources}, has the added benefit of including less-studied languages and language families\footnote{Language families were obtained  from Ethnologue: Languages of the World, available at \url{https://www.ethnologue.com/}.}, providing a more comprehensive view of the latest generation of multilingual capabilities of long-context LLMs.

\medskip
\noindent \textbf{Models} \quad We consider four proprietary and two open-source long-context LLMs. 

\begin{itemize}
\item \textbf{GPT-4} is a proprietary multilingual LLM from OpenAI \citep{achiam2023gpt} that has been shown to perform a wide range of tasks. We used the \emph{gpt-4-0125-preview} version, which is the latest one at the time of our experiments. It has a context window of {128K} tokens and was trained with the data until Dec 2023. We also study the recently introduced {\bf GPT-4o} model.

\item \textbf{Gemini-1.5} is another proprietary LLM withh a context window of {10M} tokens \citep{reid2024gemini}. We used the \emph{gemini 1.5 pro} version which is built on top of mixture-of-experts transformer-based architecture. 

\item \textbf{Claude-3} is yet another proprietary model released by Anthropic with a context window of length {200K} \citep{claude3} but is claimed to accept up to 1M tokens. We used \emph{claude-3-sonnet-20240229} variant of the Claude family.

\item \textbf{YaRN-Llama-2-7b} \citep{peng2023yarn} is an open-source model that extends Llama 2 model \citep{touvron2023llama} to accept a larger context window. It is available in different model sizes with varying context windows. We selected the 7B model with the maximum context window of {128K} tokens, accessed via Huggingface\footnote{\url{https://huggingface.co/NousResearch/Yarn-Llama-2-7b-128k}}. 

\item \textbf{Llama-3-8B} is a robust open-source model \citep{dubey2024llama3herdmodels}. We selected the instruction-tuned version of the model with a context window of 8k\footnote{\url{https://huggingface.co/meta-llama/Meta-Llama-3-8B-Instruct}}.


\end{itemize}

\subsection{Retrieval and Reasoning Tasks}

Language models with the ability to handle long context rely heavily on their capacity to retrieve relevant information from the given text and reason based on that information to interpret and follow human instructions effectively. Although synthetic tasks alone may not provide a comprehensive assessment of a language model's long-context capabilities, they offer the advantage of being easily adaptable to specific scenarios and languages. This is particularly important given that the most recent long-context real-world benchmarks are limited to English \citep{an2023eval} or bilingual English/Chinese \citep{bai2023longbench,qiu2024clongeval,yuan2024lveval}. Moreover, there is some evidence to suggest that performance on synthetic retrieval tasks can, to a certain extent, generalize to real-world datasets \citep{qiu2024clongeval}. Therefore, carefully designed synthetic tasks can serve as a valuable tool for evaluating a language model's long-context capabilities across a diverse range of languages.


The ``needle in a haystack'' task \citep{gkamradt}, similarly to the passkey retrieval task  \citep{mohtashami2023landmark,chen2023extending,liu2024lost}, 
evaluates a model's ability to extract relevant information from lengthy documents. Typically, a target sentence (the ``needle") is inserted into a corpus of documents (the context or ``haystack"), followed by a question designed to retrieve the fact in the needle. As the input text grows longer, this task typically becomes increasingly challenging.

We can formalize the problem of needles in a haystack as follows: Given the needle $n$, a context (or haystack) $c$, and a question $q$, the model is expected to generate an answer $a$. Usually, $n$, $q$, and $a$ are short, while $c$ represents a long sequence of text that can span thousands of tokens. The task can involve either a single needle $n=1$ or multiple needles $n>1$. With one or more needles, we can create {\bf retrieval} tasks, whereas with multiple needles we can construct {\bf reasoning} tasks that require the model to draw connections between different pieces of information.

\subsubsection{Retrieving a Needle ($n=1$)}
In this task, the model's objective is to locate and extract information from a single target sentence hidden somewhere in the haystack. We adopt the same needle pattern as used in previous studies  \citep{arizeai,reid2024gemini, claude3}, which takes the form: ``\textit{The special magic \{city\} number is: \{number\}}''. Here, \{city\} is randomly chosen from a list of 69 unique cities from around the world, and \{number\} is a randomly generated 7-digit number. The list of cities were automatically translated and then post-edited into all the languages. 

In English, this yields needle sentences such as ``\textit{The special magic Paris number is: 2243738}'' or in Indonesian, ``\textit{Nomor ajaib khusus untuk kota Sydney adalah 9347172}''.

The needle is then placed at different depths within the context. We experiment with five depth positions: 0\% (near the beginning), around the 25\% mark, 50\% (in the middle), about 75\% of the way through, and 100\% (towards the end of the context). The needle is placed after the first complete sentence at each specified depth to ensure a linguistically meaningful position. Finally, the model is  asked to retrieve some information (e.g., the magic number or the city) found in the needle. It is worth noting that all languages in this study use the same Hindu-Arabic numeral system.

\subsubsection{Reasoning over Multiple Needles ($n>1$)}
In real-world applications, tasks often require not just accurate text retrieval but also the ability to reason with the recalled information. To increase the challenge, we introduce a setup where multiple needles are placed within the context, requiring the model to track and reason about these different pieces of information.

The needle format remains similar to the one used in retrieval task. We discretize the positions of the target needles into four intervals: near the top (0-25\%), in the middle (25-50\% and 50-75\%), and closer to the end (75-100\%) of the context. For instance, in the 25-50\% bucket, the first needle is placed around the 25\% depth and the remaining needles are randomly placed between somewhere within the 25-50\% depth. We explore two variations of this task, with $n=2$ and $n=3$. Finally, the model is asked to generate a response based on the information (e.g., the larger or the largest magic number, or the city with the larger/largest number) found in the needles.

\begin{figure}
    \centering
    \includegraphics[width=0.3\textwidth]{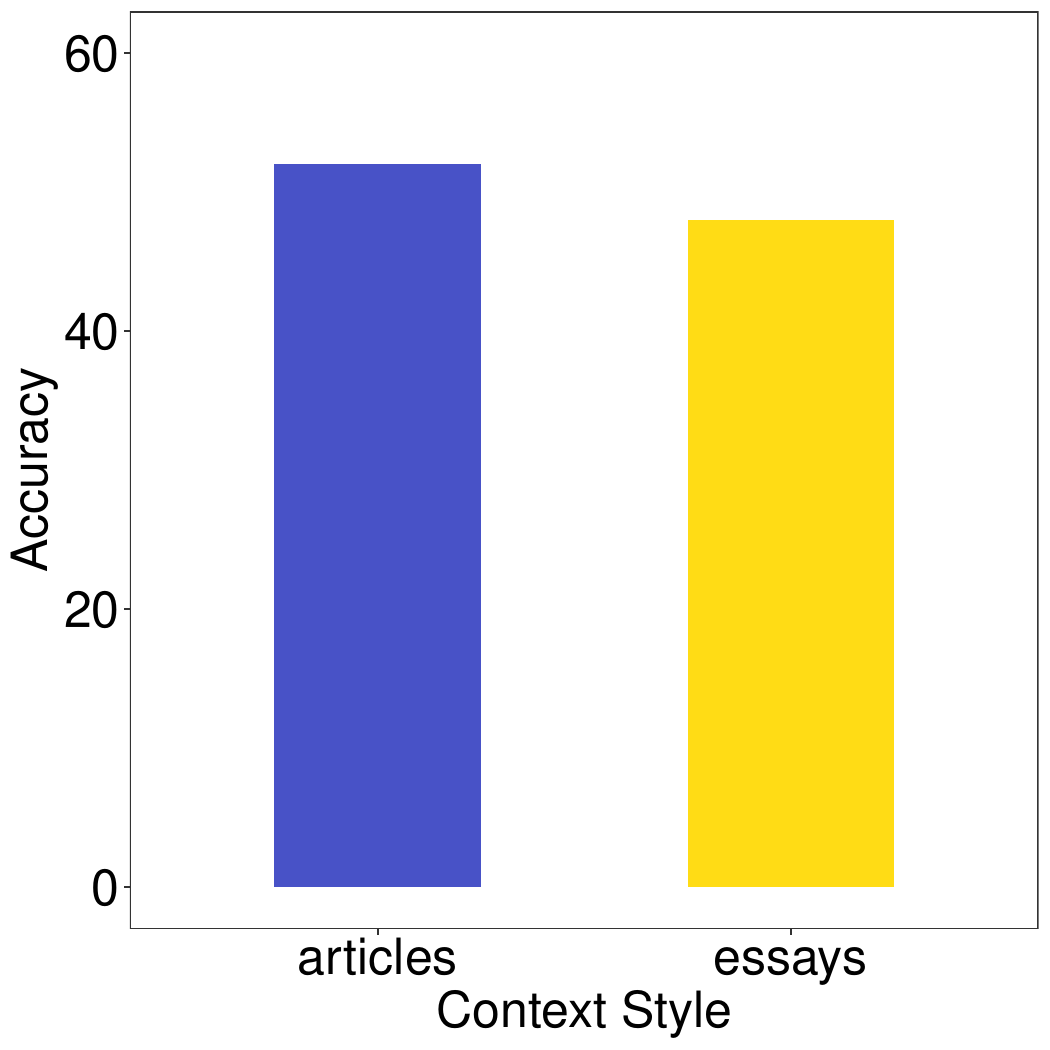}
    \caption{Ablation results of comparing Paul Graham's essays and news articles serving as haystacks for English experiments tested using GPT-4 model.}
    \label{fig:p_essays_articles}
    
\end{figure}

\subsection{Creating \texttt{mLongRR} Dataset}
Prior work  has extensively relied on Paul Graham's essays in English to make up the haystacks \citep{gkamradt, arizeai, claude3}. Translating these essays into multiple languages, however, could potentially introduce translation errors. Instead, we create a new dataset -- \texttt{mLongRR} -- by collecting BBC news articles published in the five languages\footnote{Our language selection was primarily constrained by the availability of authentic texts, especially those within the same script but varying in resource levels.}, inspired by recent work \cite{nezhad2024exploringmazemultilingualmodeling}. This approach allows us to work with sufficiently long, naturally occurring, with the added benefit that this recently published data is less likely to have been encountered by the models during training. It is worth noting that this does not result in a parallel dataset as the news articles are often specific to their respective regions. 

We assess the impacts of different data for haystacks by conducting an ablation study using the GPT-4 model. As shown in Figure~\ref{fig:p_essays_articles}, we observed no noticeable differences in the model performance when the haystacks consisted of Paul Graham's essays or news articles in English. To statistically confirm this observation, we applied McNemar’s test \cite{mcnemar1947note}, which yielded a p-value of 1.0, indicating no significant difference between the two datasets.


The haystacks in \texttt{mLongRR} were created by drawing on sufficient numbers of articles to fill up the target context window length. There are enough articles in each language to provide non-repeating text for all window lengths. Thus, for example, the first half of the 8K haystack is the same as the 4K haystack, but its second half is composed of different articles. An example of an English haystack of 8K tokens with the needle ``{\em The special magic Doha number is 9121372.}" located at 50\% depth is shown below:

\begin{Verbatim}[commandchars=\\\{\}]
    Star dunes - or pyramid dunes - are 
    named after their distinctive...
    \textit{[continues to about 4K tokens]}
    \textcolor{blue}{The special magic Doha number is}
    \textcolor{blue}{9121372.} 
    In our dark laboratory, we see light
    from these sand grains... 
    \textit{[context continues until 8K tokens]}
\end{Verbatim}

\subsection{Prompts}

In our initial run of experiments, we explored two existing prompt templates used in previous work:

\begin{itemize}
\item {\bf prompt 1} \citep{gkamradt, arizeai, claude3}, and 

\item {\bf prompt 2} \citep{reid2024gemini}.
\end{itemize}

\begin{figure}[h]
    \centering
    \includegraphics[width=0.4\textwidth]{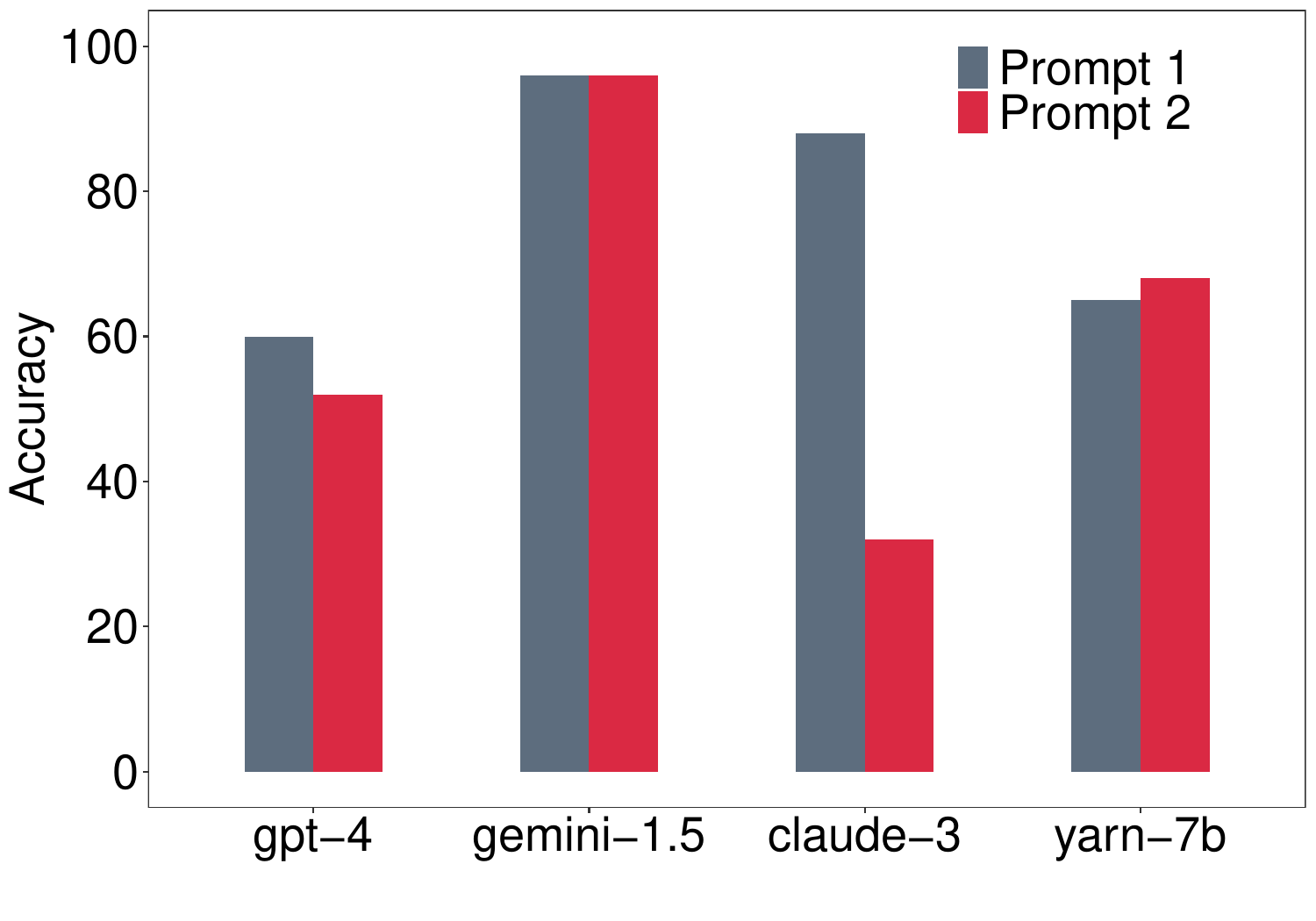}
    \caption{Ablation results of comparing two different prompts.}
    \label{fig:p_anthropic_gemini}
    
\end{figure}

As shown in Figure~\ref{fig:p_anthropic_gemini}, prompt 1 performed better in most of the models, so we decided to continue with it for the rest of the experiments. The prompt for the $n=1$ retrieval task is as follows:


\begin{verbatim}
You are a helpful AI bot that answers 
questions for a user. Keep your response 
short and direct. The following is a 
set of context and a question that will 
relate to the context. 
#CONTEXT
{context}
#ENDCONTEXT

#QUESTION
What is the special magic number? Don't 
give information outside the document or 
repeat your findings. If the information 
is not available in the context respond 
UNANSWERABLE.
\end{verbatim}



For the reasoning with multiple needles task, the prompt format remains largely similar except the question. Recall that each needle has two random variables: the number and the city. Therefore, we explore two possible reasoning questions, one about the larger number, and the other about the city with the larger number. 

\begin{itemize}
    \item For reasoning about numbers, the question is either ``{\em What is the larger magic number?}'' when  $n=2$, or ``{\em What is the largest magic number?}'' when $n>2$. 
    \item For reasoning over cities, the question is either ``{\em Which city has the larger magic number?}'' when $n=2$, or ``{\em Which city has the largest magic number?}'' when $n>2$.
\end{itemize}

Although the articles in \texttt{mLongRR} are already in the native languages, we hired four professional human translators from the freelancing website \texttt{fiverr.com} to translate the needles, city names, and prompts from English into Vietnamese, Indonesian, Swahili, and Somali. In our preliminary experiments with the GPT-4 model, we experimented with English and language-specific prompts, and found that English prompts worked better than language-specific prompts, in line with previous studies \citep{etxaniz2023multilingual, barreiss2024english, lai2023chatgpt}, for three of four non-English languages (with the exception of Swahili). As a result, the remaining experiments were conducted using English prompts.



\begin{figure*}[t]
    \centering
\includegraphics[width=1\textwidth, trim=30 40 30 0, clip]{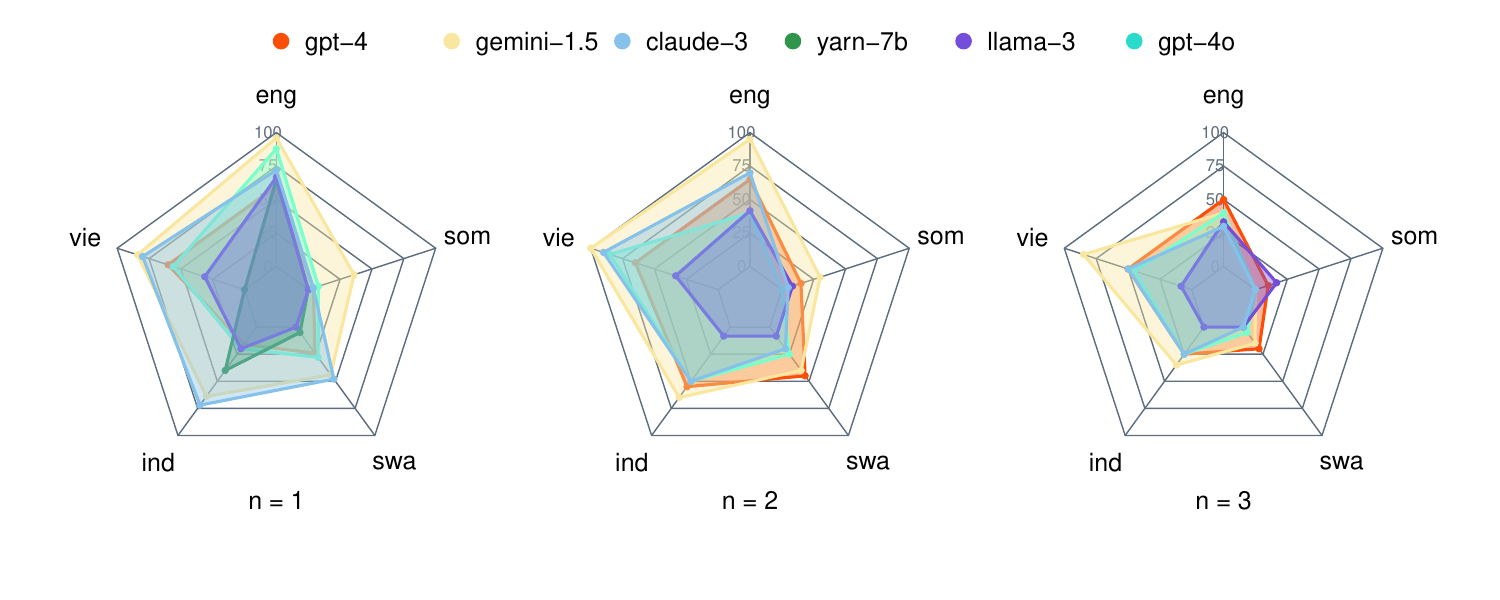}
    \caption{Radar plots showing the performance of six language models (GPT-4, Gemini-1.5, Claude-3, Yarn-7b, Llama-3, GPT-4o) across five languages (English, Vietnamese, Indonesian, Swahili, Somali) in retrieval and reasoning tasks involving one, two, and three target sentences (``needles''). The three plots represent different task complexities: single needle retrieval ($n = 1$, left plot), two needle reasoning ($n = 2$, center plot), and three needle reasoning ($n = 3$, right plot). }\label{fig:p_radar_1needle_2needle_eng_prompt}
    
\end{figure*}

\subsection{Experiments}
We conducted experiments on six models: GPT-4, Gemini-1.5, Claude-3, Yarn-7b, Llama 3, and GPT-4o. The context lengths varied from 2$k$, 8$k$, 16$k$, 32$k$, to 64$k$ tokens, and the needles were placed at different depths/positions: 0\%, 25\%, 50\%, 75\%, and 100\%. For the retrieval task, we experiment with one needle ($n=1$), whereas for the reasoning tasks, we investigate setups of needles $n=2$, and $n=3$. To enhance the robustness of our evaluation, we used a diverse corpus of recently published news articles and a combination of random cities and random numbers resulting in a vast number of possible needle variations. Furthermore, we conducted multiple runs for a subset of our experiments and consistently observed a variance close to 0 across these runs. Each model was evaluated using its default configuration, and the maximum output token size was set to 50.



\subsection{Evaluation}
The responses generated by the models were used to calculate the accuracy. For both the retrieval and reasoning tasks, the models generated a short, straightforward text containing the 7-digit number (for number-based reasoning) \cite{arizeai} or the city name (for city-based reasoning). For example, a typical output looked like this: "\texttt{3210496}" or "\texttt{The larger magic number is 8134445}". We extracted the number/city and compared it to the ground truth to check whether the model's response was correct or not.  For languages other than English, the models occasionally responded with the city name in English or the target language and both were acceptable.

\section{Results and Discussion}
This section presents our results of four main experiments: (1) performance of different models, (2) performance with respect to varying needle depths and haystack lengths, (3) performance across five different languages, and (4) reasoning over magic numbers and world cities. For the first three experiments, we analyze the models' responses when asked to retrieve and reason about the magic number. In the last experiment, we compare the models' performance when asked about the magic number or the city.

\subsection{Performance of different models across languages and tasks}

Figure~\ref{fig:p_radar_1needle_2needle_eng_prompt} presents the radar plots summarizing the the average accuracy of each model for all tasks and languages.

\begin{figure*}[t]
    \centering
    \includegraphics[width=\textwidth]{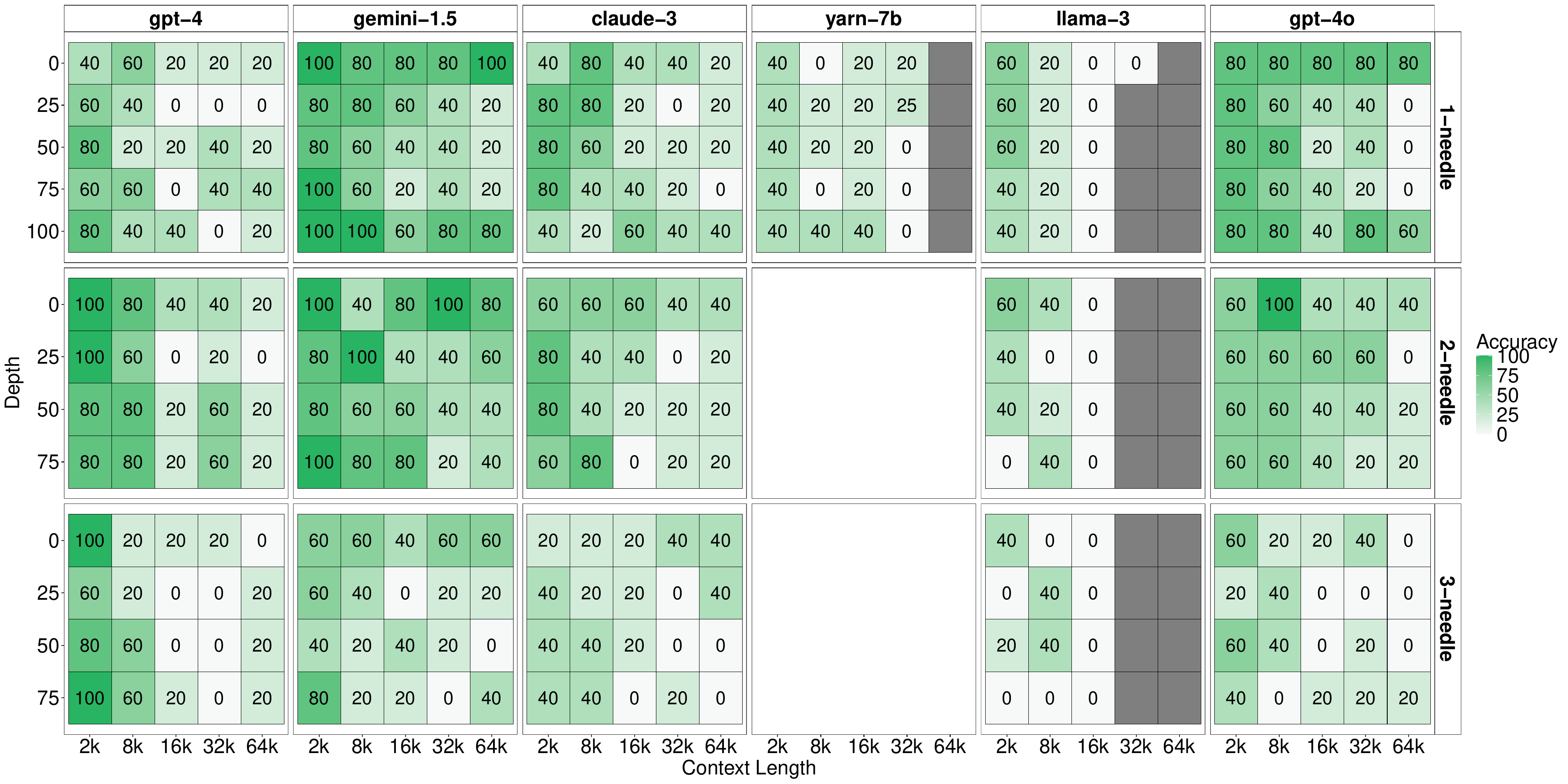}
    \caption{Heatmap visualizations with varying depths on the $y$-axis and context lengths on the $x$-axis, showing average model performance over all the languages for both retrieval (top panel) and reasoning tasks (middle and bottom panels). The color gradient from white to dark green represents accuracy levels, with darker green indicating higher accuracy.}
    \label{fig:p_model_accuracy_across_langs_eng_prompt_1_needle}
\end{figure*}

Across the languages evaluated, English generally demonstrates strong performance across all models and tasks, particularly in the simpler retrieval task ($n = 1$), likely due to the extensive amount of English data available for model training. Vietnamese also performs relatively well, especially in the more complex reasoning tasks ($n = 2$ and $n = 3$), which may be attributed to effective tokenization (more discussion in section~\ref{sec:languages}). In contrast, performance drops significantly for Indonesian, Swahili, and Somali, particularly as task complexity increases. While this decline is not surprising and highlights the ongoing challenge in multilingual NLP models trained predominantly on high-resource languages tending to perform well in those languages but faltering in low-resource languages, the extent of the decline remains noteworthy.

Gemini-1.5 and GPT-4o exhibit strong performance across all tasks and languages, maintaining the most balanced results overall, particularly in English and Vietnamese. However, their performance declines in more complex tasks for low-resource languages like Swahili and Somali. In contrast, other models display more variability, with certain strengths in specific languages but generally lower performance in reasoning tasks, particularly when multiple needles are involved.

As task complexity increases (from $n = 1$ to $n = 3$), extending from retrieval to reasoning, all models experience a performance drop, particularly in low-resource languages. This indicates that while models can handle simple retrieval tasks reasonably well, they struggle significantly with reasoning tasks that require understanding and processing long contexts in less-resourced languages.


\begin{figure*}[t]
    \centering
    \includegraphics[width=\textwidth]{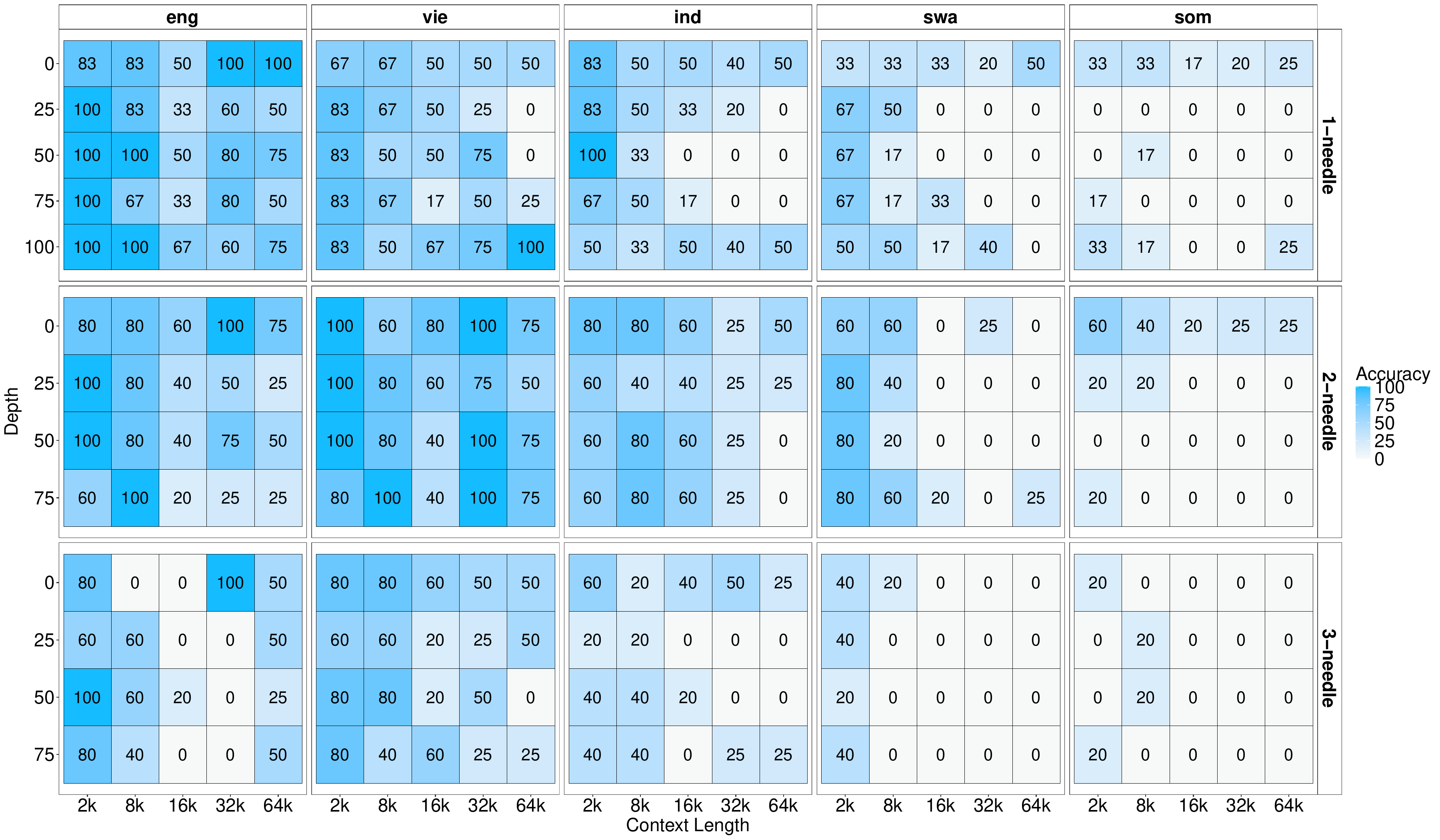}
    \caption{Language-specific heatmap visualizations with varying depths on the $y$-axis and context lengths on the $x$-axis, averaged over all the models, when $n=1$, $n=2$, and $n=3$.}
    \label{fig:p_lang_accuracy_across_model_eng_prompt_1_needle}

\end{figure*}
\vspace{1cm}

\subsection{Performance across  varying needle depths and haystack lengths}

Figure~\ref{fig:p_model_accuracy_across_langs_eng_prompt_1_needle} presents a detailed heatmap analysis of each model's performance with varying context lengths and needle depths. For all models, performance is better in shorter contexts, or when the needle is either near the top or the bottom of the context, suggesting that the ``lost in the middle'' phenomenon which was previously observed in English settings \citep{liu2024lost} extends to multilingual contexts as well. 

The heatmaps clearly show that longer context lengths and greater depths negatively impact accuracy. This suggests that current LLM architectures struggle to use relevant information effectively when processing large amounts of data or when reasoning requires multiple steps. As the task complexity increases (from retrieval to 3-needle reasoning), model performance declines across the board. This decline is particularly pronounced in models like Yarn-7b and Llama-3, which fail to handle the increased cognitive load of deeper reasoning tasks with longer contexts. Gemini-1.5 is the most resilient model across all tasks, maintaining relatively high accuracy even in complex scenarios. However, its performance also suffers as depth and context length increase, highlighting the challenges of scaling reasoning abilities in LLMs.

\setlength{\tabcolsep}{3pt}
\begin{table*}[t]
\centering
\begin{tabular}{@{}lllllll@{}}
\toprule
& \textbf{GPT-4}    & \textbf{Gemini-1.5}  & \textbf{Claude-3}  &  \textbf{YaRN-7b} & \textbf{Llama-3} & \textbf{GPT-4o}   \\ \midrule
English    & \textcolor{blue}{1.13} &\textcolor{blue}{1.15} & \textcolor{blue}{1.15} &  \textcolor{blue}{1.32} & \textcolor{blue}{1.13} & \textcolor{blue}{1.11} \\
Vietnamese & 2.08 &1.20  & \textcolor{red}{2.89} &  \textcolor{red}{2.75} & 1.27 & 1.29 \\
Indonesian & 1.92 &1.40 & 2.33 &  2.48 & 1.91 & 1.55 \\
Swahili    & 2.23 &1.85 & 2.36 &  2.48 & 2.21 & 1.68 \\
Somali     & \textcolor{red}{2.37} &\textcolor{red}{2.09} & 2.47 &  2.70 & \textcolor{red}{2.36} & \textcolor{red}{1.79} \\
\midrule
\textbf{Average} & {1.94} & {1.53} & {2.24} & {2.34} & {1.77} & {1.48} \\
\bottomrule
\end{tabular}
\caption{Tokenization rate for each language using different model tokenizers.}\label{tab:fertility}
\end{table*}

\subsection{Performance across different languages}\label{sec:languages}

The results presented in Figure~\ref{fig:p_lang_accuracy_across_model_eng_prompt_1_needle} provide a fine-grained analysis of language-specific performance. English consistently performs well across tasks, with near 100\% accuracy in simpler tasks but declining with increased complexity, particularly at greater depths and longer contexts. Vietnamese also maintains high accuracy, though it declines similarly with complexity. Indonesian starts reasonably well but drops significantly in more complex scenarios. Swahili shows weaker overall performance, struggling with all tasks, especially complex ones. Somali performs the poorest, often reaching zero accuracy as task complexity increases, highlighting challenges in handling this low-resource language. In short, performance degrades progressively as we move from high-resource languages to low-resource languages. The detailed results of each model and language are included in Appendix~\ref{app:detailed}. 

The strong performance of English and, to a lesser extent, Vietnamese, reflects the availability of ample training data in these languages. However, access to the exact language distributions of training data are not readily available for most models, including open-source model like Llama-3.

\begin{figure*}[t]
    \centering
    \includegraphics[width=0.55\textwidth]{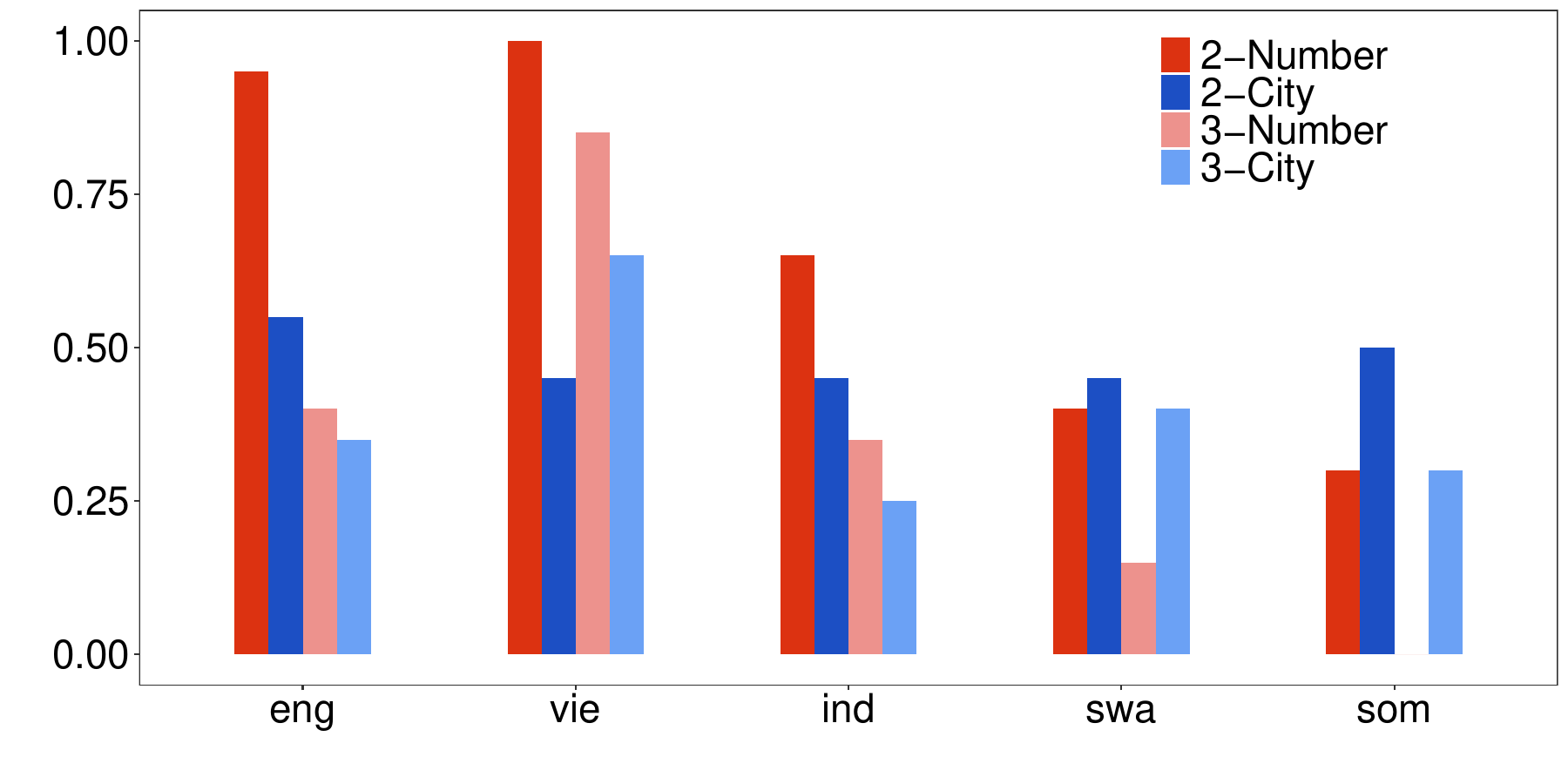}
    \caption{Comparing reasoning over magic numbers and random cities, when $n=2$ and $n=3$ (results obtained using Gemini-1.5).}
    \label{fig:p_number_city}

\end{figure*}

\subsection{Impact of tokenization}
We further analyze the tokenization  rate, also known as fertility rate, which is the average number of tokens generated per word for the different languages. The results are presented in Table~\ref{tab:fertility}. Unsurprisingly, English consistently shows the lowest tokenization rates across all models. Vietnamese has varying rates, with Gemini-1.5 being the most efficient, while Claude-3 and YaRN-7b tokenize more heavily. Indonesian exhibits moderate rates with some variability across models. Swahili and Somali, the two lowest resource level languages in our study, generally have higher tokenization rates, suggesting these are more challenging for models to process effectively.

We can make two interesting observations: (i) the performance of LLMs is influenced by the way the models tokenized text across languages, with lower fragmentation leading to improved performance, and (ii) the models with overall lower fragmentation scores, such as Gemini-1.5 followed by GPT-4o, achieved better results across all languages and tasks.

\subsection{Reasoning about magic numbers and world cities}
Lastly, we compare the performance of the models in reasoning tasks with 2 and 3 needles for two types of question prompts: identifying the larger/largest magic number (e.g., 4281932) or the city with the larger/largest magic number (e.g., Doha). From the results presented in Figure~\ref{fig:p_number_city}, we observe that the models yield generally better performance in the ``number'' tasks compared to the ``city'' tasks implying that they may be more adept at handling numerical reasoning than reasoning over geographic entities, however, this trend is reversed for Swahili and Somali.

\section{Conclusion}

We introduce a new dataset designed to study long-context retrieval and reasoning tasks across multiple languages. By evaluating six LLMs on their ability to process text in five languages with varying resource levels, using naturally occurring text and a needle-in-a-haystack paradigm with different numbers of needles, we discovered key insights. Notably, we observed a significant decline in performance, particularly when dealing with longer contexts, an increased number of needles, or lower resource levels. Even seemingly simple synthetic tasks like needle-in-a-haystack revealed substantial performance disparities. Our findings highlight the need to develop not only more effective long-context models but also improved tokenization schemes for the effective processing of low-resource languages. 



\section*{Limitations}
{While our current focus has been on languages that use Latin script, we are eager to expand our horizons and explore the diversity of languages from other scripts in the future. Furthermore, our investigation was restricted to three needles. It would be interesting to explore whether addition of more needles continues to increase the task complexity.}

\section*{Ethics Statement}
We did not implement any filtering of the haystack data, it is possible that there are inherent biases towards certain groups within the dataset. The impact of such biases on our findings remains unclear and fall outside the scope of this study. For annotation in Vietnamese, Indonesian, Swahili, and Somali, we hired translators and paid USD 15 to each translator as the short translation tasks took less than one hour each.


\section*{Acknowledgments}
We thank the anonymous reviewers as well as the
members of PortNLP lab for their insightful comments that helped improve this paper. This research
was partially supported by the National Science Foundation
grant HNDS-R 2242205.

\bibliography{main}

\appendix

\section{Detailed model and language-specific results}\label{app:detailed}
\begin{figure*}[!t]
    \centering
    \includegraphics[width=0.8\textwidth]{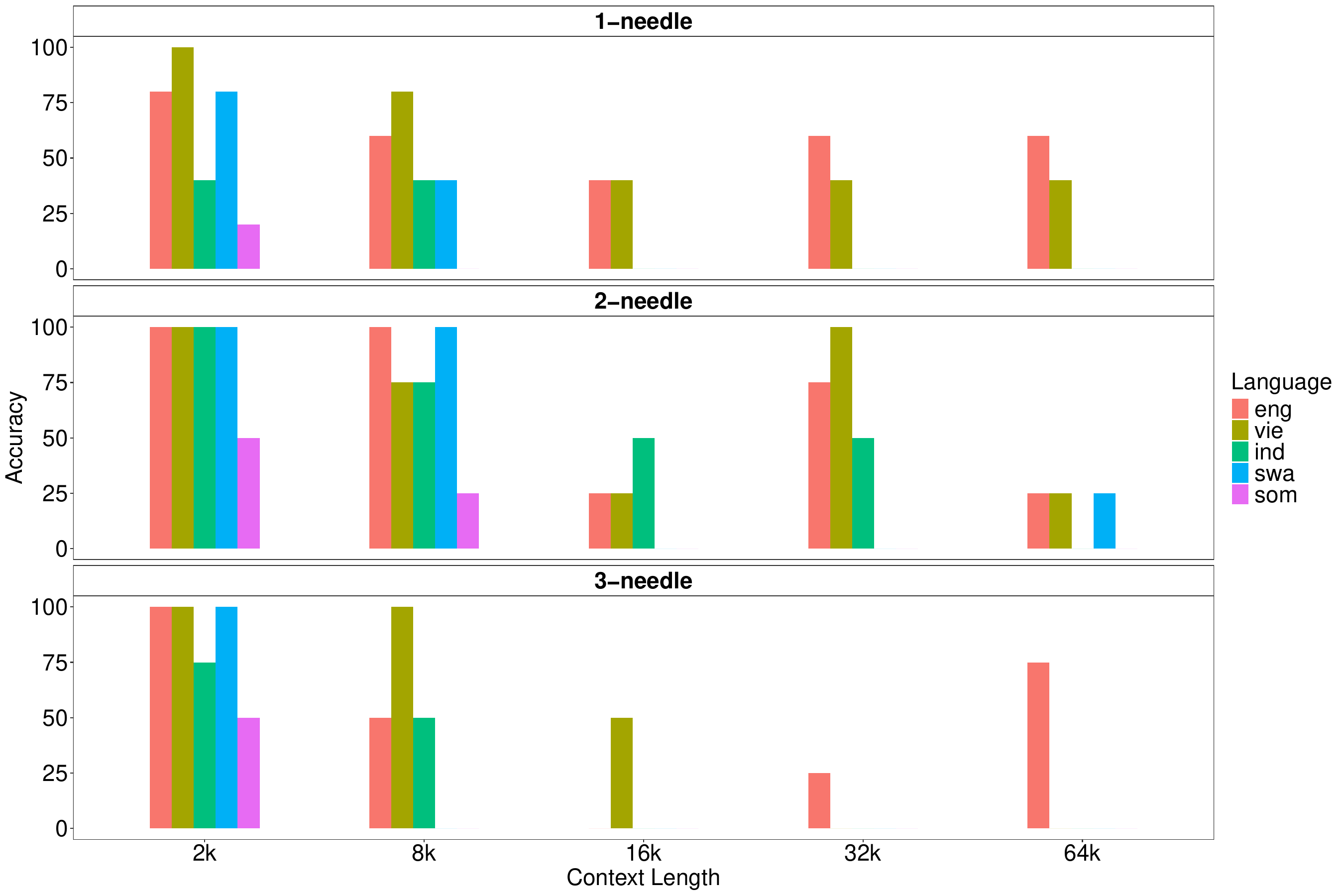}
    \caption{GPT-4}
    \label{fig:p_gpt4_performance}
\end{figure*}

\begin{figure*}[!t]
    \centering
    \includegraphics[width=0.8\textwidth]{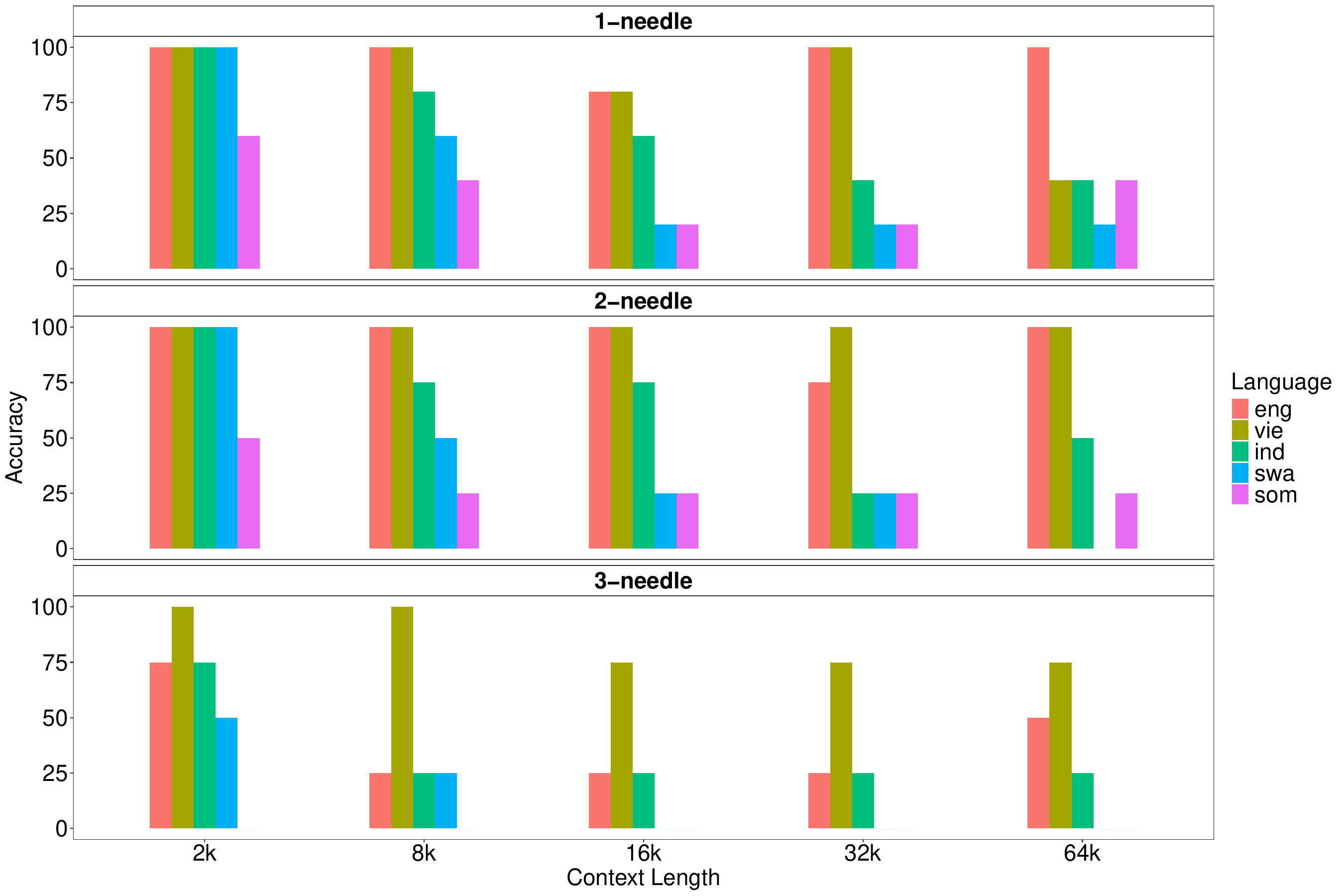}
    \caption{Gemini-1.5}
    \label{fig:p_gemini_1_5_performance}
\end{figure*}

\begin{figure*}[!t]
    \centering
    \includegraphics[width=0.8\textwidth]{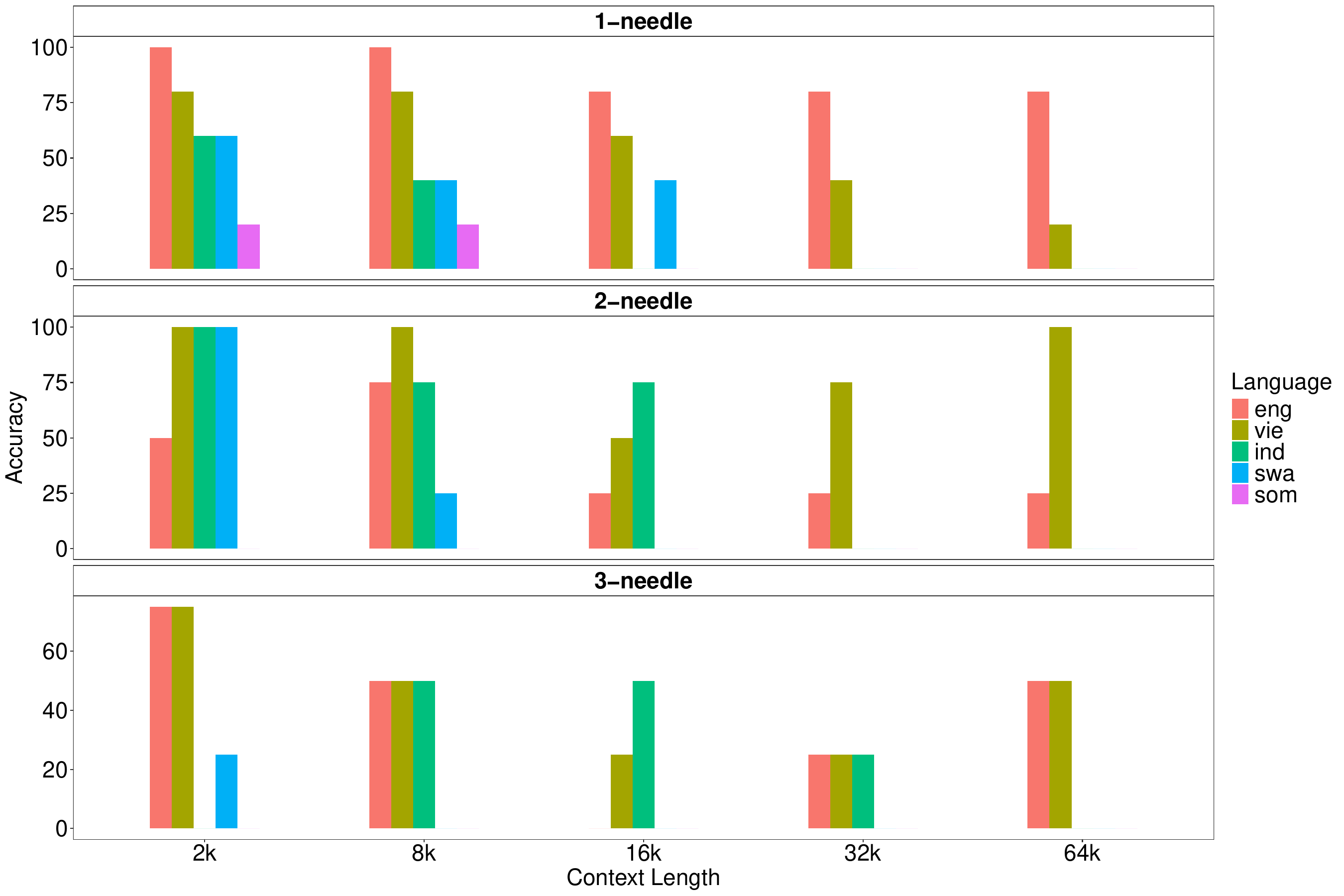}
    \caption{Claude-3}
    \label{fig:p_claude_3_performance}
\end{figure*}

\begin{figure*}[!t]
    \centering
    \includegraphics[width=0.8\textwidth]{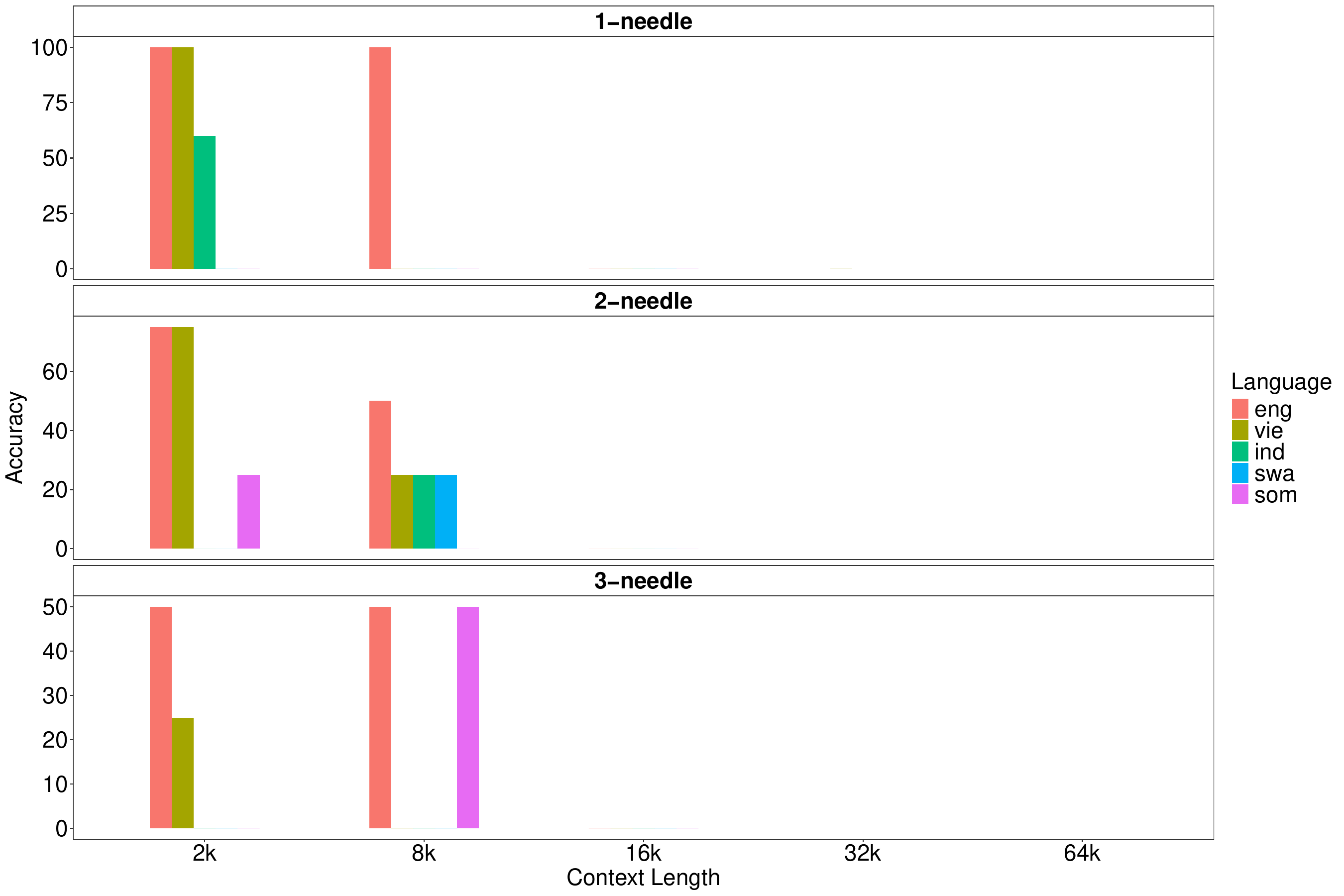}
    \caption{Llama 3}
    \label{fig:p_llama-3_performance}
\end{figure*}

\begin{figure*}[!t]
    \centering
    \includegraphics[width=0.8\textwidth]{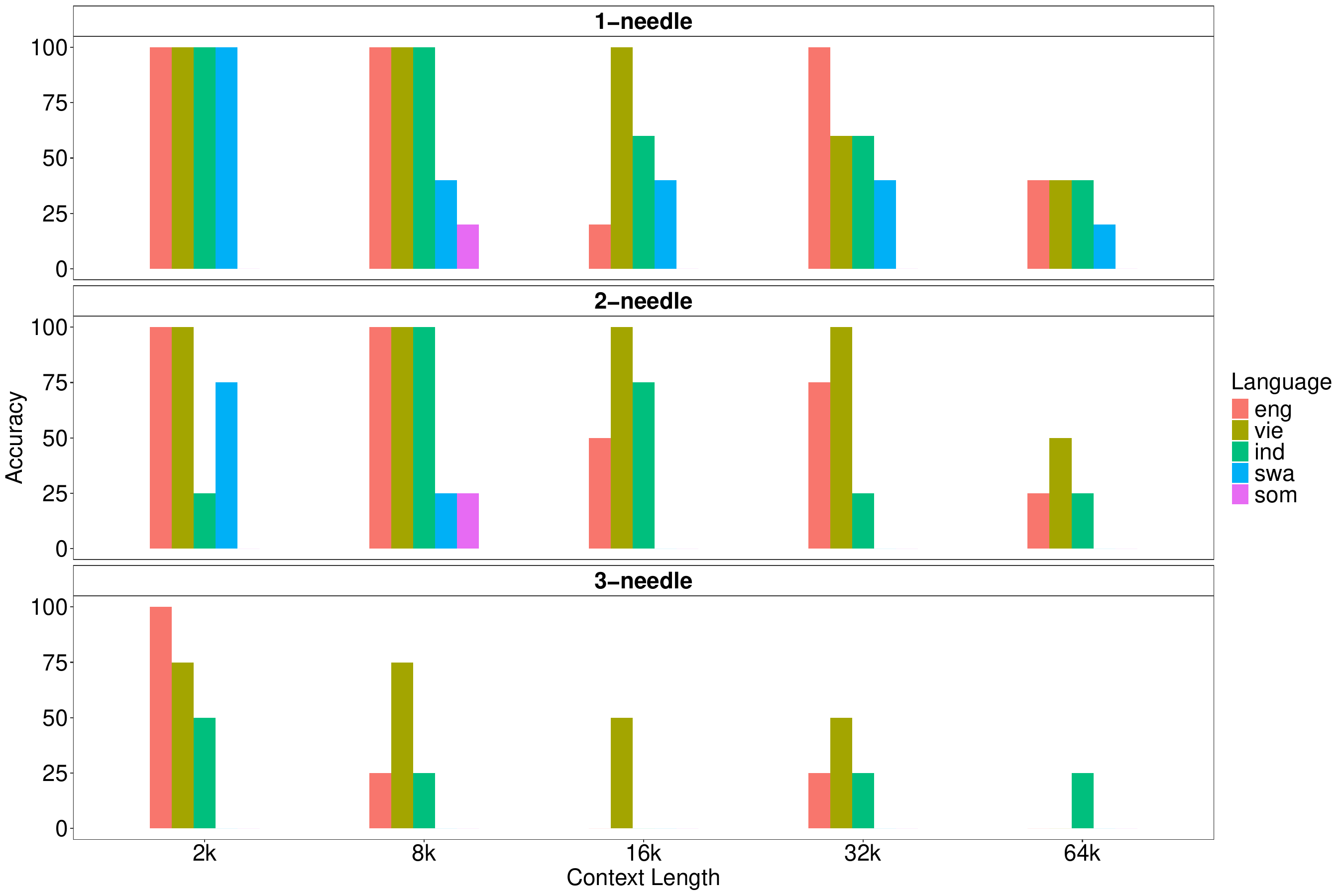}
    \caption{GPT-4o}
    \label{fig:p_gpt-4o_performance}
\end{figure*}

Figures~\ref{fig:p_gpt4_performance}, \ref{fig:p_gemini_1_5_performance}, \ref{fig:p_claude_3_performance}, \ref{fig:p_llama-3_performance} and \ref{fig:p_gpt-4o_performance} show detailed results of the five models: GPT-4, Gemini-1.5, and Claude-3, Llama 3, and GPT-4o.

\end{document}